\documentclass[10pt]{article}

\usepackage[T1]{fontenc}
\usepackage{lmodern}

\usepackage[super,sort&compress]{natbib}

\usepackage{graphicx}
\usepackage{float}
\usepackage{multirow}
\usepackage{booktabs}
\usepackage{amsmath}
\usepackage{amsfonts}
\usepackage{xspace}

\usepackage[letterpaper, margin=1in]{geometry}

\usepackage[T1]{fontenc}
\usepackage{lmodern}

\usepackage{caption}
\begin{document}

\bibliographystyle{unsrt}

\newcommand{\ourmethod}{{BOxCrete}\xspace}

\begin{center}
{\LARGE \textbf{\ourmethod: A Bayesian Optimization Open-Source AI Model for Concrete
Strength Forecasting and Mix Optimization}}\\[1.5em]

{\large Bayezid Baten$^{1\dagger}$, M. Ayyan Iqbal$^{1\dagger}$,
Sebastian Ament$^{2}$, Julius Kusuma$^{2}$ and Nishant Garg$^{1*}$}\\[1.5em]

$^{1}$Department of Civil and Environmental Engineering,
University of Illinois Urbana-Champaign, Urbana, IL, USA\\
$^{2}$Meta,
Menlo Park,
California, USA.\\[1.5em]

\textit{\dag\ The authors contributed equally}

\vspace{10pt}

\textit{*Corresponding author email: nishantg@illinois.edu}\\
\end{center}

\begin{abstract}
Modern concrete must simultaneously satisfy evolving demands for mechanical
performance, workability, durability, and sustainability, making mix designs
increasingly complex.
Recent studies leveraging Artificial Intelligence (AI) and Machine Learning
(ML) models show promise for predicting compressive strength and guiding mix
optimization, but most existing efforts are based on proprietary industrial datasets and
closed-source implementations.
Here we introduce \ourmethod, an open-source probabilistic modeling and optimization
framework trained on a new open-access dataset of over 500 strength measurements
(1--15~ksi) from 123 mixtures---69 mortar and 54 concrete mixes tested at five curing
ages (1, 3, 5, 14, and 28~days).
\ourmethod leverages Gaussian Process (GP) regression to predict strength development,
achieving average R$^2 = 0.94$ and RMSE $= 0.69$~ksi,
quantify uncertainty, and carry out multi-objective optimization of compressive strength and embodied carbon.
The dataset and model establish a reproducible open-source foundation for data-driven development of AI-based optimized mix designs.
\end{abstract}

\newpage

\section*{Introduction}
Concrete is the second most widely consumed material on Earth, after water,
with an estimated annual production of approximately 10 billion m$^3$
(\textasciitilde 30 billion tonnes)\cite{Habert2020decarbonization}.
Its most expensive and CO$_2$ intensive ingredient, cement, is currently
produced at the rate of 4 billion tonnes/year\cite{Krausmann2018metabolism}.
With the global population forecasted to increase by 2.5 billion by 2050, the
societal need for housing and public infrastructure will further drive the
production of cement and concrete\cite{Swilling2018urbanization}.
Considering the scale and complexity of concrete ingredients, efficient mixture
proportioning is a key strategy for optimizing the use of our planet's
resources.
Modern concrete mixes are increasingly expected to simultaneously achieve
several objectives, including but not limited to: a) meet structural
performance requirements (early and late age strengths), b) meet specified
durability criteria (low permeability and volumetric stability), c) meet
workability requirements (be pumpable and placeable on site), d) be
cost-effective, and e) have minimal carbon footprint\cite{DeRousseau2018optimization}.
Over the last decade, increased interest in minimizing carbon footprint has
made modern concrete mixes even more complex.

Specifically, given that 1 tonne of Portland cement contributes \textasciitilde
0.8 tonne of CO$_2$ during its production and is globally responsible for
6--8\% of anthropogenic CO$_2$ emissions, there has been significant interest
in partially or fully replacing this key ingredient\cite{Scrivener2018ecoefficient,Ostovari2021mineralization,Witte2024gwp,Olsson2023pathways}.
Among these approaches, partial replacement of cement with Supplementary Cementitious Materials
(SCMs), such as Fly Ash (FA) and Ground Granulated Blast Furnace Slag (GGBFS),
is one of the most practical and feasible strategies\cite{Snellings2023scm,Gartner2015binder}.
In addition, as traditional SCMs become limited in supply, other SCMs, such as
calcined clays, harvested ashes, and natural pozzolans, are receiving increased
attention worldwide\cite{Min2025calcined,Juenger2011alternative,Scrivener2018lc3}.
Most of these SCMs not only reduce the carbon footprint and cost of concrete
but also contribute to long-term performance benefits, including enhanced
late-age strength, durability, and lower permeability, thereby extending the
service life of concrete infrastructure\cite{Juenger2015scmrole,Lothenbach2011scm}.

However, there are two core issues with high SCM usage.
First, early-age strength and workability can suffer depending on the
quality and quantity of SCM used\cite{Juenger2019sources}.
Additionally, modern concrete mixes containing blended cements tend to
incorporate increased quantities of multiple organic admixtures to mitigate
challenges arising from novel ingredients\cite{Li2021superplasticizers,Plank2015admixtures}.
The compressive strength trajectory of these modern concrete mixes thus becomes
a highly nonlinear function of multiple mix parameters\cite{Young2019strength,Flatt2023blended,Kosmatka2011design,Mindess2003concrete}.
This high-dimensional interdependence has rendered traditional concrete mix
design largely empirical, relying on a time- and resource-intensive
trial-and-error approach\cite{DeRousseau2018optimization}. Concrete production is inherently regionalized, as it depends strongly on the availability, variability, and cost of locally sourced binders, admixtures, aggregates, and water. Consequently, universal mix prescriptions are infeasible, and formulations must be systematically adapted to local material variability and project-specific performance targets
As a result, there is a growing need for data-driven mix design frameworks that
can intelligently navigate this complex material space and guide the
development of sustainable, low-cost concrete mixes that meet structural,
durability, and workability requirements on-site.

Machine Learning (ML) techniques have been increasingly explored to address
this challenge, offering the potential to model complex nonlinear interactions
and predict performance metrics directly from mix
compositions\cite{Yeh1998ann,Li2022ml,Nunez2021compressive,Asteris2021hybrid,Song2022flyash}.
Prior studies have applied ML models to both curated lab datasets and
industrial ready-mix records to predict compressive strengths.
Based on the current literature, there are two major research gaps.
First, the original dataset from Yeh (1998)\cite{Yeh1998ann}, consisting of
\textasciitilde 1000 measurements, is nearly 30 years old, and since then there
has been no new publicly available dataset obtained at a single laboratory.
Recent studies on large industrial datasets (\textasciitilde 10k measurements
by Young et~al.\cite{Young2019strength} and \textasciitilde 40k measurements by Pfeiffer
et~al.\cite{Pfeiffer2024bayesian}) are promising but not publicly accessible.
Second, most existing approaches neglect the probabilistic and time-evolving
nature of strength development and fail to quantify uncertainty, both essential
for robust structural applications where early and late strengths are equally
important\cite{Pfeiffer2024bayesian,Ament2023sustainable}.
To address these gaps, there is a critical need for publicly available,
high-fidelity datasets and open-source AI models that can forecast strength
over time and quantify the associated uncertainty.

\begin{figure}[]
    \centering
    \includegraphics[width=0.8\textwidth]{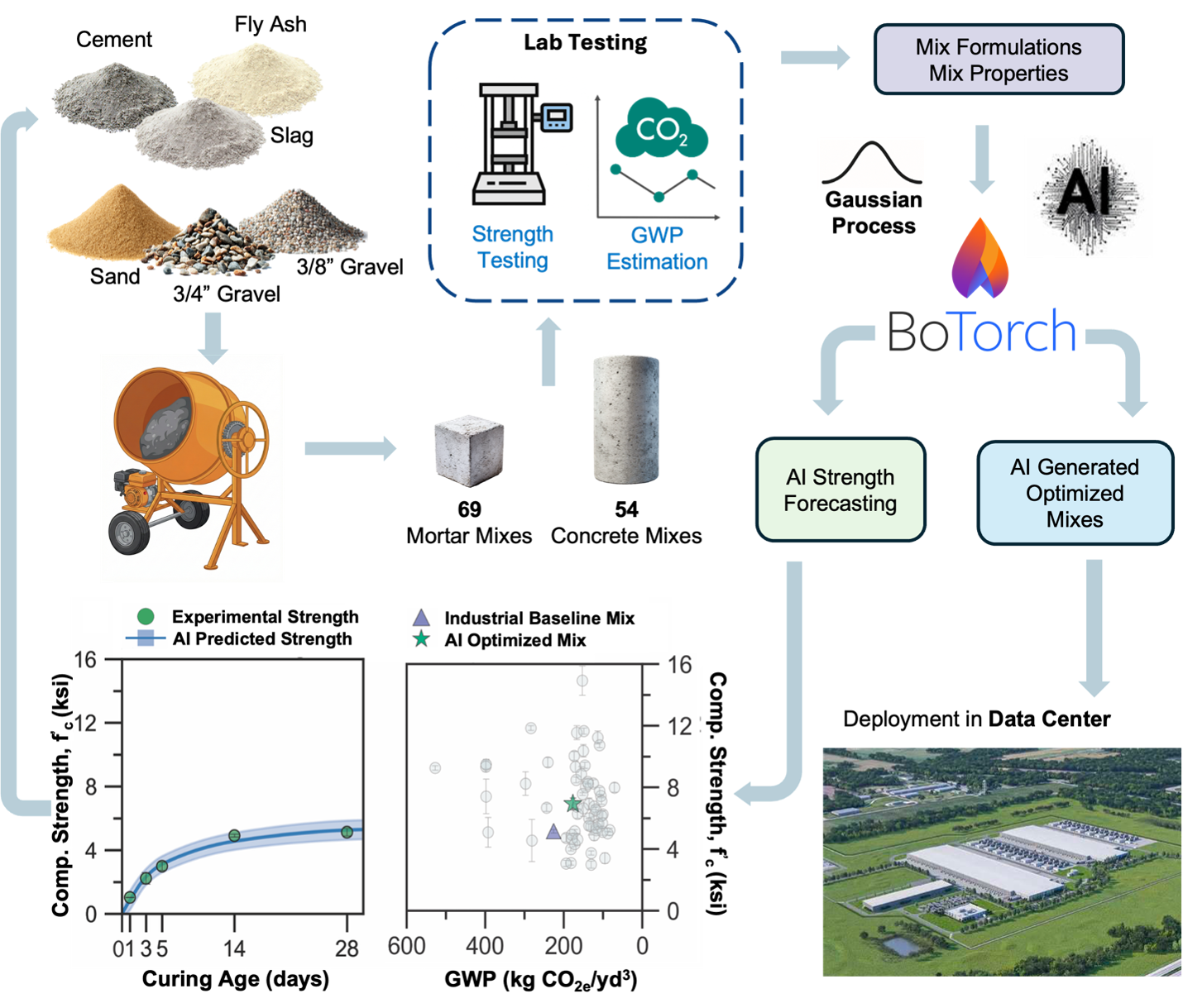}
    \caption{
\textbf{AI-in-the-loop framework of \ourmethod for sustainable concrete mix optimization.}
Raw materials (Type IL cement, SCMs, and aggregates) are proportioned into mortar and concrete mixes, cast, and tested for compressive strength at 1, 3, 5, 14, and 28 days. Cradle-to-gate Global Warming Potential (GWP) is quantified via life cycle assessment (LCA). The resulting dataset of mix proportions, strength, and GWP is used to train a Gaussian Process Regression model to forecast strength evolution with quantified uncertainty. Coupled with Bayesian Optimization (BoTorch), the trained model generates new mix designs targeting user-defined strength and sustainability objectives. AI-suggested formulations are experimentally validated and reintroduced into the loop, enabling adaptive learning and accelerated discovery of high-performance, sustainable concrete for infrastructure applications.
}
    \label{fig:Fig1}
\end{figure}

In this study, we introduce \ourmethod—a novel, open-source modeling and optimization
framework trained on 533 unique strength measurements, which we
release as an open-access public dataset.
These 533 measurements correspond to 123 experimentally cast mortar (69)
and concrete (54) mixes produced at a
single laboratory (University of Illinois) with systematically varied
water-to-binder ratios, binder contents, and SCM replacement levels (As shown in Fig.~\ref{fig:Fig1}).
\ourmethod AI model leverages Gaussian Process (GP) regression\cite{Rasmussen2004gp} to probabilistically model strength development as a
function of time and mix composition, 
enabling uncertainty-aware predictions
across multiple curing ages\cite{Ament2023sustainable}.
To accelerate material design, we couple this
framework with Bayesian Optimization (BO)\cite{frazier2018tutorial} to perform
multi-objective optimization\cite{Daulton2020ehvi, Ament2023logei} 
of compressive strength and Global Warming
Potential (GWP), enabling rapid identification of optimal binder formulations.
The model was iteratively refined across six experimental phases
(Phase I--VI), each incorporating increasingly diverse data, and was validated
against 12 independent concrete mixes excluded from training. The final model
achieves an R$^2$ of 0.94 and RMSE $<$ 0.70 ksi across all curing ages,
demonstrating high predictive accuracy and strong generalization. Finally,
\ourmethod also enabled the development of concrete mixes with compressive
strengths exceeding 6~ksi at 28~days while maintaining embodied carbon levels
as low as 120--150~kg~CO$_2$e/yd$^3$, emphasizing its capability to
simultaneously optimize performance and sustainability over the long term. The
complete dataset and AI model are being released as open-access resources to
enable community-driven refinement and broader applicability. The combined open-source AI framework and curated experimental dataset reduce barriers to AI adoption while enabling accelerated discovery of high-performance, low-carbon concrete formulations across diverse material systems. By coupling
experimental precision with statistically interpretable machine learning,
\ourmethod establishes a scalable framework for multi-objective optimization
in concrete mix design, advancing the development of sustainable,
high-performance construction materials.

\section*{Results and Discussion}
\textbf{Strength and Sustainability of Experimental Mixes.}
\vspace{10pt}

Understanding the coupling between mechanical performance and embodied carbon
(sustainability) is essential for advancing sustainable binder design
strategies.
Previous studies have shown that optimized utilization of Supplementary
Cementitious Materials (SCMs), such as fly ash and slag, enables a balance
between strength development and CO$_2$ reduction through complementary
hydration mechanisms\cite{Scrivener2018ecoefficient,Skibsted2019reactivity}.
To establish a comprehensive predictive framework, it is therefore necessary
to characterize how compressive strength and Global Warming Potential (GWP)
evolve across a compositionally diverse binder space.
This forms the foundation of the present study, providing the
mechanistic and statistical basis for subsequent model development.

\begin{figure}[H]
    \centering
    \includegraphics[width=0.9\textwidth]{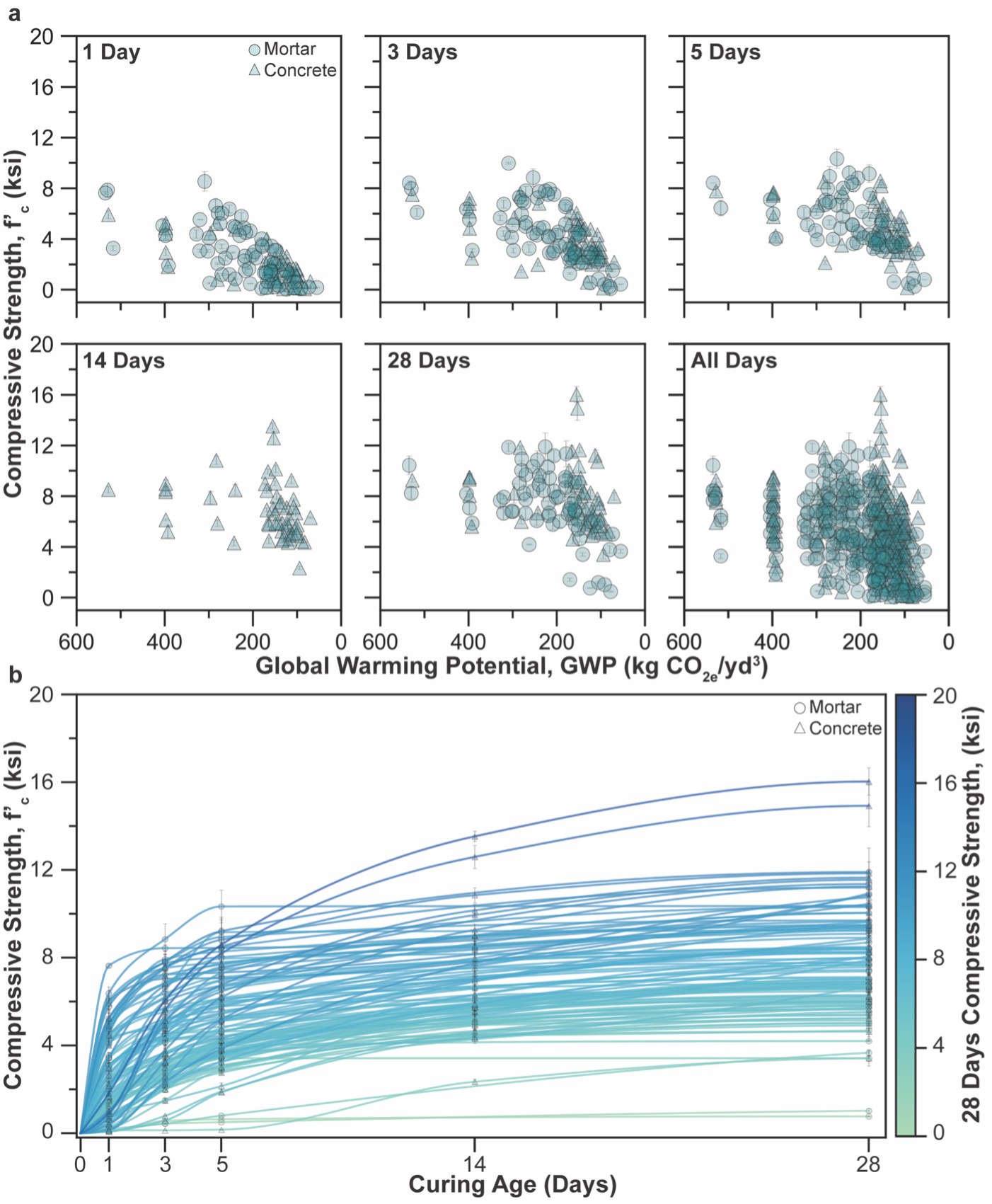}
\caption{
\textbf{Evolution of compressive strength and embodied carbon in mortar and concrete mixes across curing ages.}
Global warming potential (GWP; kg~CO$_2$e/yd$^3$) is plotted against unconfined compressive strength ($f'_c$; ksi) for 69 mortar (circles) and 54 concrete (triangles) mixes at 1, 3, 5, 14, and 28 days of curing. Each data point represents a unique mix design with varying binder content (240--2150~lb/yd$^3$), water-to-binder (w/b) ratios (0.20--0.50), and different replacement levels of Portland Limestone Cement with Class C and Class F fly ash and ground granulated blast furnace slag. Vertical error bars indicate the standard error of the compressive strength measurements based on three replicate specimens. GWP values were computed on a cradle-to-gate basis using standardized material emission factors. The optimized performance zone, defined by low embodied carbon and high compressive strength, corresponds to the top-right quadrant of each plot.
}
    \label{fig:strength_evolution}
\end{figure}

Fig.~\ref{fig:strength_evolution} presents the compressive strength and corresponding GWP values for 123
unique binder formulations, comprising 69 mortar and 54 concrete mixes.
Each subplot in Fig.~\ref{fig:strength_evolution}a corresponds to a discrete hydration age (1, 3, 5, 14,
and 28~days), illustrating the temporal progression of both mechanical
performance and embodied carbon across a broad design space.
Fig.~\ref{fig:strength_evolution}b further demonstrates the evolution of compressive strength with curing
age for all mixtures, highlighting the highly heterogeneous, nonlinear
strength-development trajectories captured within the dataset.

This open-access dataset spans a wide variation in mix proportions, including
water-to-binder ratios from 0.20--0.50, total binder contents from
240--2150~lb/yd$^3$, and varying proportions of SCMs such as fly ash and slag.
This diversity results in compressive strengths ranging from below 1~ksi to as
high as 16~ksi (6.9--110~MPa), and GWP values spanning 55--535~kg~CO$_2$e/yd$^3$.

At early hydration ages (1--5~days; Fig.~\ref{fig:strength_evolution}a), an increase in compressive
strength is generally associated with higher GWP, particularly in mixes with
elevated OPC content.
This correlation arises from the rapid hydration kinetics of alite (C$_3$S),
which drive early nucleation and growth of C--S--H and portlandite\cite{Scrivener2019hydration}.
Strength gain during this early period is dominated by primary hydration
products, and the extent of reaction strongly correlates with cement
content\cite{Xu2024molecular}.
Consequently, high early strength is typically achieved at the expense of
higher clinker usage and associated CO$_2$ emissions.
Mortar systems tend to cluster at higher GWP levels relative to concrete due to
the absence of aggregate dilution, which increases binder volume per unit
mixture.

By 28~days (Fig.~\ref{fig:strength_evolution}a), the strength--GWP relationship becomes non-monotonic.
Several mixtures with GWP values between 100 and 200~kg~CO$_2$e/yd$^3$ achieve
compressive strengths between 10 and 16~ksi.
This shift reflects the activation of SCMs, which contribute to strength
through secondary C--S--H formation driven by pozzolanic reactions (fly ash)
and latent hydraulic reactions (slag)\cite{Skibsted2019reactivity,Aughenbaugh2013flyash,Kocaba2012slag}.
These reactions progress more slowly than OPC hydration but become increasingly
significant with curing time.
As hydration proceeds, blended systems exhibit enhanced reaction degrees,
densified microstructures, and refined pore-size distributions, often
producing strengths comparable to or exceeding high-cement
controls\cite{Berodier2015pore}.
Furthermore, concrete mixtures consistently show lower GWP values than mortars
at all ages due to the dilution effect of coarse and fine aggregates that
occupy volumetric space without contributing to binder-related emissions.

Overall, the results demonstrate that strength evolution is highly sensitive to
mix composition, yet low embodied carbon can be achieved concurrently through
balanced binder design.
This highlights the need to delineate the underlying compositional design
space, emphasizing the pivotal role of individual mixture parameters in
governing both mechanical performance and sustainability.
Compared to previous studies, which typically report 28-day strengths of
3--7~ksi and GWP values of 240--450~kg~CO$_2$e/yd$^3$, the present dataset
covers a substantially broader domain—capturing strengths up to 16~ksi and GWP
values ranging from approximately 55 to 535~kg~CO$_2$e/yd$^3$.

Across hydration ages, the dataset reveals an increasing density of
high-performance, low-carbon mixtures within the upper-right quadrant of the
strength--GWP space, defined here as the optimization target.
These mixtures reflect an effective balance of reduced clinker content and
enhanced SCM reactivity.
The dynamic evolution of strength--GWP profiles across curing time underscores
the importance of time-dependent mix optimization and the potential role of
machine-learning frameworks in guiding prescriptive design of sustainable,
high-performance concrete mixtures.

\vspace{10pt}

\noindent\textbf{Progressive learning of temporal strength evolution in
concrete.}
\vspace{10pt}

Predicting the temporal evolution of compressive strength in cementitious
systems requires models that can accurately represent the nonlinear coupling
between hydration kinetics, binder chemistry, and volumetric mixture
proportions.
Traditional empirical and neural-network-based approaches have demonstrated
usefulness for predicting single-age strengths\cite{Young2019strength,Yeh1998ann}, but they
often lack generalizability and interpretability when extended to continuous
strength-development trajectories.
To address these limitations, the present study employs a Gaussian Process (GP) regression framework capable of capturing time-dependent hydration
behavior while quantifying predictive uncertainty.
This section evaluates the phase-wise learning progression of the model and its
ability to reproduce experimental strength profiles across diverse mixture
designs.

In this study, the AI model was progressively trained and refined through
sequential enhancement of the dataset across six mix-development phases
(Phase~I to Phase~VI).
The final, fully trained model (Phase~VI) was then used to predict compressive
strength–development curves for a held-out set of 12 concrete mixtures.
This validation set was randomly selected and not included in the training
data.
To further assess model robustness, five distinct testing sets (Testing
Sets~1--5) were constructed over five random iterations.

The compositional variation of the 12 concrete mixtures in these testing sets
is graphically summarized in Fig.~S3 of the Supplementary Information.
Fig.~\ref{fig:evolution_of_strength} shows the progression of the model's predictive capability for six
representative concrete mixtures (Panel~A), comparing predicted compressive
strength values with experimental measurements over curing ages of 1–28~days
across the training phases (Phase~I--VI) for Testing Set~1.
These phases correspond to progressively larger training datasets, ranging from
a small initial subset (Phase~I) to the complete concrete dataset (Phase~VI),
and were used to incrementally calibrate the AI model (see Table~2).
Predicted strength curves (blue lines) include 95\% confidence intervals
(shaded regions, $\pm 2$ standard deviations), while experimental measurements
are plotted as discrete points with error bars representing $\pm 1$ standard
deviation ($n=3$).
The corresponding comparisons for the remaining mixtures in Testing Set~1 and
for Testing Sets~2--5 are shown in Fig.~S4 of the Supplementary Information.

The model successfully reproduces the characteristic sigmoidal
strength-evolution behavior exhibited across all mixtures, reliably capturing
the multiphase kinetics of cement hydration.
In particular, the model captures both the rapid early-age strength gain
observed between 1–5~days and the gradual, steady increase in strength at later
ages (14–28~days).
Because experimental measurements were collected at multiple key points during
strength development (1, 3, 5, 14, and 28~days), the GP regression model is able to learn
the nuances associated with early primary hydration and later-stage SCM-driven
reactions.
This strong agreement across diverse mixture compositions for all testing sets
(Fig.~\ref{fig:evolution_of_strength} and Fig.~S4) demonstrates the model's ability to infer underlying
hydration kinetics directly from compositional and temporal data without
relying on empirical curve-fitting parameters.

\begin{figure}[]
    \centering
    \includegraphics[width=1\textwidth]{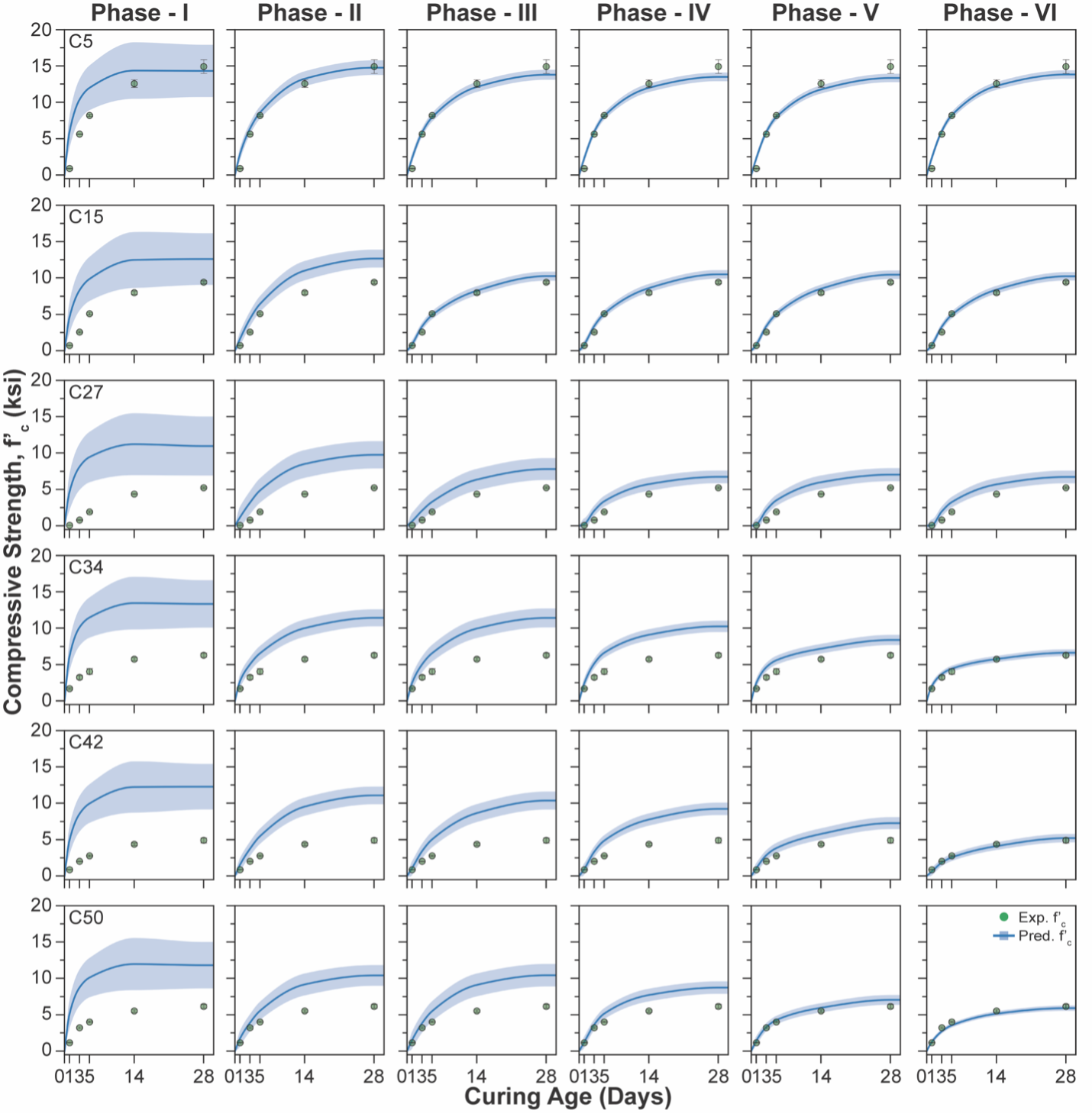}
    \caption{
\textbf{Evolution of AI model predictions for compressive strength as a function of training dataset size.}
Predicted (blue lines) and experimentally measured (green points) compressive strength ($f'_c$, ksi) are shown for six representative concrete mixes across six training dataset phases (Phase I to Phase VI). Each phase corresponds to an expanded training set size, incorporating an increasing number of mix designs with varying binder compositions and water-to-binder (w/b) ratios. Shaded regions indicate the 95\% confidence interval (mean $\pm 2$ standard deviations) of model predictions. Experimental data points represent the mean $\pm 1$ standard deviation from triplicate tests ($n = 3$). The figure illustrates progressive improvement in model accuracy and uncertainty reduction as the training set becomes more comprehensive.
}
    \label{fig:evolution_of_strength}
\end{figure}

The model accuracy improves systematically with the size and compositional
diversity of the training dataset, as reflected by the progressive
contraction of the shaded uncertainty bounds across phases in Fig.~\ref{fig:evolution_of_strength}.
In the early learning stages (Phase~I--II), the prediction intervals are
comparatively wide (approximately $\pm$0.8--2.4~ksi at 28~days), particularly
for mixtures characterized by distinct binder chemistries or elevated
water-to-binder ratios.
This spread indicates higher predictive uncertainty due to the model's limited
exposure to diverse hydration pathways.
As additional formulations are incorporated during Phases~III--V, the shaded
uncertainty regions narrow substantially to $\pm$0.4--1.9~ksi, and the
predicted mean values show improved alignment with experimental observations
across all curing ages.
By Phase~VI, the uncertainty envelope collapses to within approximately
$\pm$0.2--1.3~ksi for most data points, with predicted and measured strengths
demonstrating near-unity correlation (R$^2 \approx 0.94$, RMSE $\approx$
0.73~ksi).

Notably, the model captures not only overall strength trends but also strength
differences arising from variation in mixture parameters.
For example, Mix~C5, with a lower w/b ratio of 0.25, reaches 1-day and 3-day
strengths of approximately 2.2~ksi and 6~ksi, respectively, whereas Mix~C15,
with a higher w/b ratio of 0.35 (see Supplementary Mix Design Excel file),
exhibits lower early-age strengths of roughly 0.71~ksi and 3.2~ksi.
Increasing data density and compositional diversity enable the model to better
learn nonlinear hydration interactions, reducing bias at intermediate ages
(3--14~days), where earlier phases exhibited mild underestimation.
Importantly, this monotonic improvement and convergence pattern is consistently
observed across all five independent testing iterations (Sets~1--5),
underscoring the reproducibility and robustness of the sequential learning
framework.

Compared to previous AI-based models—which typically report R$^2$ values
between 0.49 and 0.89 and RMSE values near 0.75~ksi\cite{Young2019strength,Yeh1998ann,Pfeiffer2024bayesian}
(see Table~S11)—the present GP regression framework attains
substantially higher accuracy (R$^2 \approx 0.94$, RMSE $<$ 0.69~ksi) while
maintaining interpretability through its probabilistic formulation and explicit
uncertainty quantification on an open-access dataset.
The phase-wise learning strategy enables efficient characterization of
nonlinear hydration behavior, allowing the model to progressively distinguish
clinker-dominated early-age reactions from SCM-driven pozzolanic contributions
that govern later-age strength evolution.
The adaptive contraction of uncertainty bounds with dataset enrichment
quantitatively reflects improved epistemic confidence and generalization
across curing ages.
Overall, the refined model provides a statistically robust and interpretable
basis for temporal strength prediction across diverse binder chemistries,
forming a critical foundation for data-driven optimization of sustainable
concrete systems.

\vspace{10pt}

\noindent\textbf{Phase-wise convergence of accuracy and uncertainty in
predictive strength modeling}

\vspace{10pt}

\begin{figure}[]
    \centering
    \includegraphics[width=0.9\textwidth]{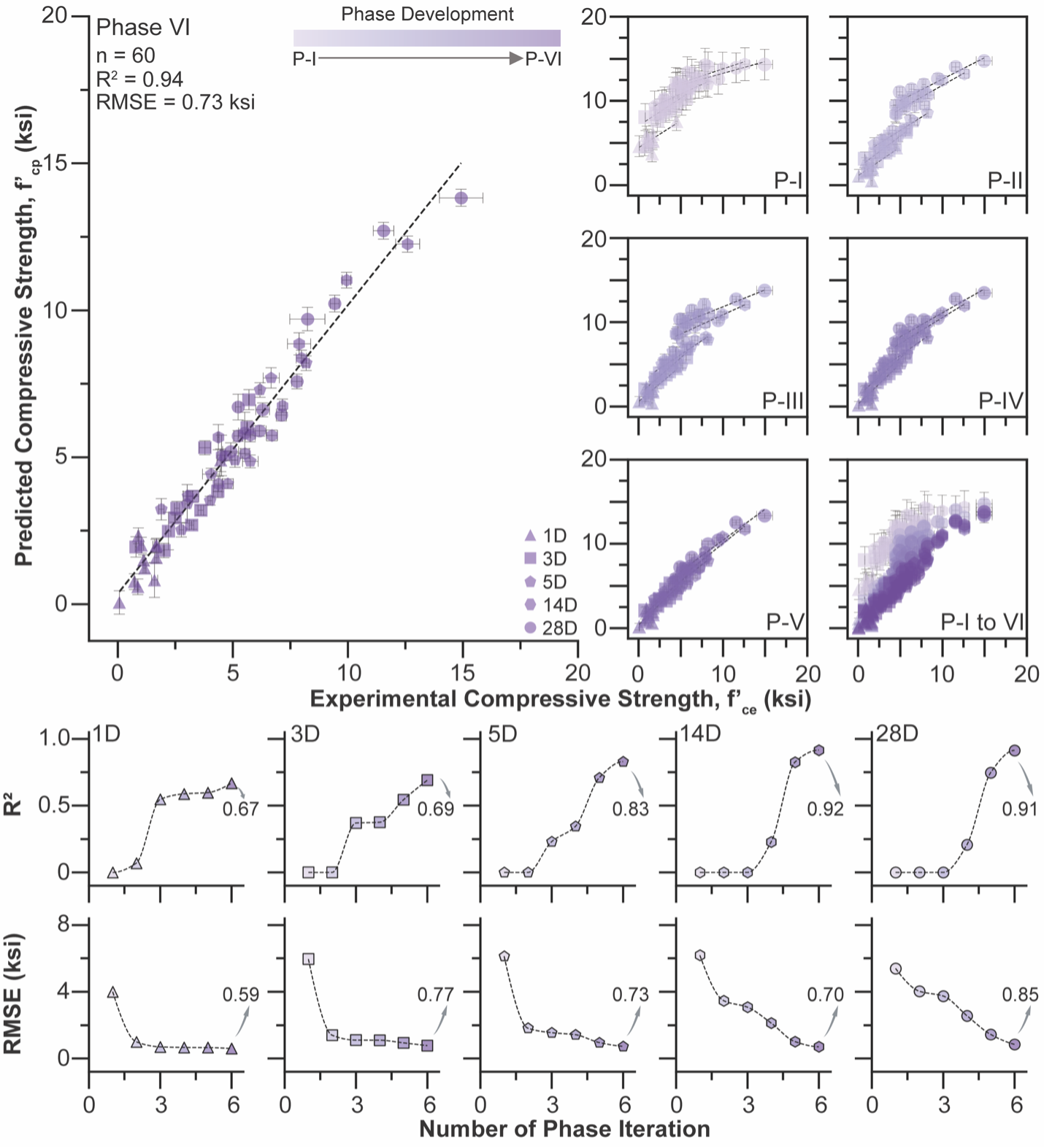}
   \caption{
\textbf{Phase-wise improvement in \ourmethod predictive accuracy for Testing Set 1.}
Model performance is shown across six training phases (P-I to P-VI) using compressive-strength data from 123 mortar and concrete mixes evaluated at 1, 3, 5, 14, and 28 days.
\textbf{Left:} Final-phase (P-VI) parity plot comparing predicted ($f'_{c,p}$) and experimental ($f'_{c,e}$) strengths for $n = 60$ independent test samples ($R^2 = 0.94$, RMSE $= 0.73$~ksi). Marker shapes denote curing age; error bars represent $\pm 1$ standard deviation (total span $= 2\sigma$). Strong alignment with the 1:1 line indicates high predictive accuracy across hydration stages.
\textbf{Right:} Phase-specific parity plots (P-I to P-VI) and aggregated panel (P-I to VI) demonstrate progressive reduction in scatter and bias, with early phases underestimating high-strength mixes and later phases converging closely to parity.
\textbf{Bottom:} Evolution of $R^2$ and RMSE with phase iteration for each curing age shows consistent improvement, with the largest gains at 14 and 28 days due to improved representation of SCM-driven delayed hydration reactions.
}
\label{fig:phase_improvement}
\end{figure}

The predictive capability of the \ourmethod model was further evaluated through
a systematic phase-wise learning framework designed to quantify how accuracy
and uncertainty evolve with sequential data enrichment.
This approach provides a statistical interpretation of model convergence
behavior, linking the incorporation of compositional diversity to measurable
improvements in prediction accuracy across curing ages.

Fig.~\ref{fig:phase_improvement} presents the progressive evolution of the \ourmethod model's
predictive accuracy across six sequential training phases (P--I to P--VI),
demonstrating how performance improves with the inclusion of increasingly
diverse mixture compositions.
The main parity plot (left) compares the experimentally measured and predicted
compressive strengths ($f'_c$) for all twelve concrete mixtures at the final
training phase (P--VI), achieving R$^2 = 0.94$ and RMSE $= 0.73$~ksi.
Marker shapes correspond to curing ages (1, 3, 5, 14, and 28~days), while the
color gradient denotes phase progression.
The strong clustering of points along the 1:1 line and the narrow scatter
envelope reflect excellent predictive fidelity and generalization across a
wide strength range.

The phase-wise parity plots (right) capture the transition in predictive
performance from P--I to P--VI.
In the early phases (P--I and P--II), wider scatter and systematic
underestimation are evident, particularly for mixtures exceeding 6~ksi, due to
limited representation of SCM-rich or high-strength compositions.
With progressive data augmentation (Phases P--III to P--V), the scatter around
the parity line contracts, and the slope of the best-fit line approaches unity.
By Phase~VI, the shaded uncertainty bands shrink from $\pm$2.4~ksi in early
phases to approximately $\pm$0.2~ksi, and the predicted and measured strengths
align closely along the 1:1 reference, indicating near-complete elimination of
systematic bias.
The cumulative panel (P--I to P--VI) further reinforces this convergence
behavior, demonstrating that performance improves consistently across
retraining cycles.

The quantitative metrics in the lower panels strengthen these observations.
The Coefficient of Determination (R$^2$) and Root-Mean-Square Error (RMSE) are
plotted as functions of phase number for each curing age (see Tables~S6--S10).
Both metrics exhibit systematic enhancement with phase progression—R$^2$
increasing from $\sim$0 in Phase~I to $>$0.94 in Phase~VI, and RMSE
decreasing from $\sim$5.5~ksi to $<$0.73~ksi. The contraction of uncertainty
envelopes reflects a consistent reduction in epistemic error, with the model
gaining statistical confidence as it assimilates previously unseen
compositional domains. The most pronounced improvements occur at later ages (14
and 28~days), where the inclusion of SCM-rich compositions enables better
modeling of delayed pozzolanic and latent-hydraulic reactions. In contrast,
early-age predictions (1--3~days), governed primarily by alite hydration and
admixture effects, show smaller relative improvement, consistent with the lower
compositional sensitivity of early hydration processes. Similar improvements in
R$^2$ and RMSE are observed across all five testing iterations (Datasets~2--5),
as shown in Fig.~S5 of the Supplementary Information. A summary of R$^2$ and
RMSE improvements for each testing set across phases is provided in Figs.~S6
and S7. A consistent convergence of RMSE across all five testing
datasets—from initial values exceeding 3000~psi to below 1000~psi by
Phase~VI—demonstrates the model's capacity to efficiently learn
compositional–strength relationships and generalize to unseen formulations.
The converged R$^2$ and RMSE values after Phase~VI are summarized in Table~1.

The predictive accuracy achieved in this work (R$^2 = 0.94$, RMSE = 0.69~ksi)
is comparable to, and in several respects exceeds, established benchmarks for
strength-prediction models of cementitious systems.
Pfeiffer~et~al.~(2024)\cite{Pfeiffer2024bayesian} employed a Gaussian Process regression
model trained on a private dataset of 9{,}296 industrial concrete mixtures and
reported R$^2 \approx 0.88$ with RMSE $\approx$ 0.91~ksi (see Table~S11).
Notably, \ourmethod achieves similar or better accuracy with nearly two orders
of magnitude fewer data points, highlighting its data efficiency.
This improvement arises from deliberate compositional diversification and the
phase-wise training strategy, which collectively enable robust learning of
high-dimensional, nonlinear relationships between mixture parameters and
compressive strength.
Similarly, Young~et~al.~(2019)\cite{Young2019strength} analyzed over 10{,}000
field-collected mixtures and reported R$^2 \approx 0.60 \pm 0.05$ and RMSE $=$
0.64--0.72~ksi, though their dataset and models were proprietary.
While producing similar RMSE values, \ourmethod offers a key advantage: its
dataset and code are fully open-access, enabling reproducibility, transparency,
and extensibility beyond proprietary implementations.
Finally, the early ANN framework of Yeh~(1998)\cite{Yeh1998ann} achieved R$^2
\approx 0.88$ for 1{,}030 strength measurements but lacked the compositional
diversity needed for generalizable predictions of modern concrete mixtures.

Collectively, these results demonstrate that sequential learning prioritizing
data diversity over volume significantly enhances predictive precision and
interpretability. The Gaussian Process Regression framework not only captures
time-dependent strength evolution but also quantifies uncertainty reduction,
linking statistical convergence with physical hydration mechanisms. This
phase-wise learning paradigm establishes a reproducible and data-efficient
pathway for developing high-accuracy, physically interpretable AI models for
sustainable concrete design.

\begin{table}[H]
\centering
\small
\caption{Performance metrics of the \ourmethod model across testing datasets and
curing ages. Statistical performance across five independent testing sets is
shown in terms of the Coefficient of
Determination (R$^2$) and Root-Mean-Square Error (RMSE) for 1-, 3-, 5-, 14-, and 28-day curing ages. The global averages
yield R$^2$ = 0.94 ± 0.01 and RMSE = 0.69 ± 0.07 ksi.}
\begin{tabular}{l l c c c c c c}
\toprule
\textbf{Metric} & \textbf{Age} & \textbf{Set 1} & \textbf{Set 2} & \textbf{Set 3} & \textbf{Set 4} & \textbf{Set 5} & \textbf{All Sets} \\
\toprule

\multirow{6}{*}{\textbf{R$^2$}}
& 1-Day     & 0.67 & 0.94 & 0.86 & 0.86 & 0.67 & 0.80 $\pm$ 0.12 \\
& 3-Days    & 0.69 & 0.82 & 0.82 & 0.86 & 0.69 & 0.78 $\pm$ 0.08 \\
& 5-Days    & 0.83 & 0.83 & 0.86 & 0.82 & 0.83 & 0.83 $\pm$ 0.02 \\
& 14-Days   & 0.92 & 0.85 & 0.87 & 0.93 & 0.92 & 0.90 $\pm$ 0.04 \\
& 28-Days   & 0.91 & 0.84 & 0.84 & 0.93 & 0.91 & 0.89 $\pm$ 0.04 \\
& All Ages  & 0.94 & 0.94 & 0.93 & 0.94 & 0.96 & 0.94 $\pm$ 0.01 \\
\midrule

\multirow{6}{*}{\textbf{RMSE (ksi)}}
& 1-Day     & 0.59 & 0.35 & 0.55 & 0.64 & 0.59 & 0.54 $\pm$ 0.11 \\
& 3-Days    & 0.77 & 0.66 & 0.74 & 0.76 & 0.77 & 0.74 $\pm$ 0.05 \\
& 5-Days    & 0.73 & 0.70 & 0.71 & 0.96 & 0.73 & 0.77 $\pm$ 0.11 \\
& 14-Days   & 0.70 & 0.70 & 0.71 & 0.65 & 0.70 & 0.69 $\pm$ 0.02 \\
& 28-Days   & 0.85 & 0.84 & 0.84 & 0.71 & 0.85 & 0.82 $\pm$ 0.06 \\
& All Ages  & 0.73 & 0.67 & 0.72 & 0.75 & 0.58 & 0.69 $\pm$ 0.07 \\
\bottomrule
\end{tabular}

\label{tab:performance}
\end{table}

\section*{\ourmethod for Inverse Mix Design of Concrete}
\vspace{10pt}
Beyond forward prediction, the \ourmethod framework enables inverse design of
concrete mixtures based on user-defined compressive strength and Global Warming
Potential (GWP) constraints.
This capability is particularly relevant for practitioners aiming to meet
structural performance requirements while minimizing embodied carbon during
early-stage mix design.
To demonstrate this functionality, \ourmethod was used to generate candidate
mixtures that satisfy minimum compressive strength thresholds of 5000, 6000,
7000, and 8000~psi across GWP bins of 50~kg~CO$_2$e/m$^3$.
For each GWP bin, the model returned a set of 50 viable mix designs.

Fig. S8 presents violin plots showing the distributions of key mixture
parameters—including cement, fly ash, slag, Water-to-Binder (w/b) ratio, and
High-Range Water Reducer (HRWR) dosage—across different GWP intervals and
strength levels.
As strength requirements increase, the model progressively narrows the feasible
design window, particularly in terms of allowable w/b ratio and total binder
content, reflecting the stricter mechanical constraints associated with
higher strengths.
At lower GWP levels, the model favors SCM-rich formulations, especially those
incorporating higher slag contents and increased HRWR dosages, which compensate
for reduced cement fractions while maintaining adequate workability and
performance.

Fig. S9 shows the resulting compositional design space, illustrating how
binder content, w/b ratio, and HRWR dosage vary with compressive strength at 1
and 28~days.
These mixtures represent the solution space explored by the model to satisfy
user-specified strength and sustainability constraints.

Beyond pointwise strength–carbon tradeoffs, the inverse design framework allows
users to prescribe full strength-development profiles—including target S-curve
trajectories over time—and obtain corresponding mixture compositions.
By integrating temporal evolution into the design process, \ourmethod enables
performance-driven customization of concrete mixtures that meet both early-age
and long-term strength requirements under defined sustainability criteria.
This expands the applicability of the framework for tailored,
specification-driven mix design in modern low-carbon construction.

\section*{Limitations and Future Outlook}
Despite our open-source AI model's strong predictive performance (R$^2$ = 0.94,
RMSE = 0.69 ksi) on our open-access, public dataset of 500+ unique strength
measurements, this work has two key limitations.

Firstly, the current dataset primarily focuses on varying the volumetric
proportions of a limited type of ingredients.
Changing the nature, source, and quality of an ingredient (such as SCM or
chemical admixture) would significantly affect concrete performance.
The currently trained model, thus, may not work well on other materials that
differ from those used in the training sets.
That said, this open-source approach marks an initial step towards developing a
more general model that can account for the quality of various ingredients by
considering both chemical (e.g., reactivity) and physical (e.g., particle size
distributions) properties.
Given the open-access nature of our work, we invite the concrete research
community to build upon this initial dataset and strive towards a more general
model for concrete performance prediction.

Secondly, the current model primarily considers mechanical performance
(compressive strength) and sustainability (GWP) as the key optimization
parameters.
However, two additional parameters important for field
performance -- workability and durability—are not currently considered.
These were considered out of scope, given our initial focus on strength
development curves.
Future studies should consider optimizing and predicting all parameters to
fully enable robust inverse mix design for any project-specific criteria.

Despite these limitations, we are optimistic that this study opens up
a new pathway towards the development and deployment of open-source AI models
on open-access datasets for both concrete performance prediction and mix
design. In particular, by providing an open-source repository under the MIT license, 
we want to encourage the implementation of AI in commercial software that can be
used by concrete suppliers, materials suppliers, and the construction industry 
at-large to accelerate the discovery of high-performance, sustainable concrete
mixes using both standard and novel materials.

\section*{Conclusions}

Modern concrete is a complex mixture of multiple ingredients---all of which have a non-linear impact on the final mix performance. Given growing interest in optimizing and designing concrete mixes for multiple objectives such as mechanical performance, sustainability, workability, and durability, there is a need for developing AI/ML-based approaches.

This study introduces \ourmethod, an open-access machine learning framework
developed to predict and optimize the compressive strength
and environmental performance of concrete mixes with high Supplementary
Cementitious Material (SCM) contents.
Trained on a systematically curated dataset of 123 unique mixes (69 mortar and 54 concrete mixes),
\ourmethod enables multi-objective optimization of mix designs that balance
performance and sustainability.
We have four key conclusions from this study.

Firstly, the Gaussian process-based model demonstrates high predictive accuracy
across both early and late curing ages, with R$^2$ exceeding 0.90 and reaching
$>$~0.94 at 28~days along with RMSE of 0.69 ksi.
This improvement underscores the model's capability to capture the delayed
reactivity of SCMs such as fly ash and slag.

Secondly, coupling \ourmethod with Bayesian Optimization identifies
formulations exceeding 10~ksi compressive strength at 28 days while maintaining Global
Warming Potential (GWP) between 150--200~kg~CO$_2$e/m$^3$.
These optimized mixes achieve $>$~50\% cement replacement and up to 60\% lower
GWP than equivalent control mixes.

Thirdly, the model's probabilistic architecture enables interpretable,
uncertainty-aware predictions, allowing decision-makers to quantify confidence
levels, prioritize data collection, and explore strength–sustainability
trade-offs efficiently.

Finally, broader data diversity and standardized reporting are essential to
scale \ourmethod globally.
Integrating regional datasets and employing transfer learning will enhance
adaptability across variable SCM chemistries and local material supplies.

\ourmethod establishes a reproducible and scalable framework for
AI-assisted sustainable concrete design, accelerating progress toward low-carbon
infrastructure.

\section*{Experimental Methods}

\noindent\textbf{Machine Learning Framework for Time-Dependent Strength
Prediction}
\vspace{10pt}

\noindent{\textit{Gaussian Processes: Learning from Data}}
\vspace{10pt}

Gaussian Processes (GPs) are a non-parametric approach for modeling functions
based on complex and noisy prior data\cite{Rasmussen2004gp}
and has been implemented in frameworks including BoTorch~\cite{Balandat2019BoTorch} and GPyTorch~\cite{Gardner2018GPyTorch}.
Rather than returning a single deterministic outcome, a GP provides a posterior
predictive distribution over possible outputs, conditioned on observed
input–output pairs.
A GP defines a distribution over functions and is characterized by the property
that any finite collection of function evaluations follows a joint Gaussian
distribution\cite{Rasmussen2004gp} 
While a multivariate Gaussian distribution models probability distributions
over finite-dimensional vectors, a GP models distributions over
functions.

Formally, a GP is defined by a mean function $\mu = m(X)$ and a covariance
(kernel) function $\Sigma = k(X, X')$.
Given a set of input points
\[
X = [x_1, x_2, \dots, x_n],
\]
the associated function values are jointly distributed as
\begin{equation}
f \sim \mathcal{N}(\mu, \Sigma),
\label{eq:gp_prior}
\end{equation}
where
\[
f = f(X) = [f(x_1), f(x_2), \dots, f(x_n)],
\qquad
\mu = m(X) = [m(x_1), \dots, m(x_n)],
\]
and the covariance matrix is defined as
\[
\Sigma_{ij} = k(x_i, x_j').
\]
This matrix is positive semidefinite when $X = X'$.

The primary goal of GP regression is to predict unknown function values $f^{*}$
at new inputs $X^{*}$, given the observed training outputs $y$.
Assuming observational noise
\[
\varepsilon \sim \mathcal{N}(0, \sigma_n^2 I),
\]
the training observations are modeled as
\[
y = f + \varepsilon.
\]

The joint distribution of the noisy training observations $y$ and the latent
test outputs $f^*$ is given by
\begin{equation}
\begin{bmatrix}
y \\
f^{*}
\end{bmatrix}
\sim
\mathcal{N}
\left(
\begin{bmatrix}
\mu \\
\mu^{*}
\end{bmatrix},
\begin{bmatrix}
\Sigma + \sigma_n^2 I & \Sigma_{*}^{T} \\
\Sigma_{*} & \Sigma_{\*\*}
\end{bmatrix}
\right),
\label{eq:gp_joint}
\end{equation}
where
\[
\Sigma = k(X, X),\quad
\Sigma_{*} = k(X^{*}, X),\quad
\Sigma_{\*\*} = k(X^{*}, X^{*}).
\]

The conditional distribution (posterior) over $f^{*}$ given the training data
is:
\begin{equation}
f^{*} \mid y \sim \mathcal{N}\!\left(
\mu^{*} + \Sigma_{*}^{T}(\Sigma + \sigma_n^2 I)^{-1} (y - \mu),\;
\Sigma_{\*\*} - \Sigma_{*}^{T}(\Sigma + \sigma_n^2 I)^{-1}\Sigma_{*}
\right).
\label{eq:gp_post}
\end{equation}

For a dataset $D = (X, y)$, the posterior GP can be written as
\begin{equation}
f \mid D \sim \mathcal{GP}(m_D, k_D),
\end{equation}
with posterior mean
\begin{equation}
m_D(x)
=
m(x)
+
k(X, x)^{T} (\Sigma + \sigma_n^2 I)^{-1} (y - m(X)),
\label{eq:gp_post_mean}
\end{equation}
and posterior covariance
\begin{equation}
k_D(x, x')
=
k(x, x')
-
k(X, x)^{T}(\Sigma + \sigma_n^2 I)^{-1}k(X, x').
\label{eq:gp_post_cov}
\end{equation}

The posterior variance is always reduced relative to the prior variance because
it incorporates observed data, thereby reducing epistemic uncertainty.

\vspace{10pt}

\noindent{\textit{Multi-Objective Bayesian Optimization}}
\vspace{10pt}

With a GP model trained and validated, it becomes possible to identify input
variables that achieve desired output properties using Bayesian Optimization
(BO)\cite{frazier2018tutorial}, with 
applications in 
aerospace engineering~\cite{lam2018advances},
biology and medicine~\cite{pharma},
civil engineering~\cite{Ament2023sustainable},
machine learning hyperparameter optimization~\cite{snoek2012practical, turner2021bayesian}, 
and materials science~\cite{ament2021sara}.
A GP–BO framework provides a probabilistic estimate of achieving target
outcomes, thereby reducing the number of required experimental trials.
In real-world applications, many design tasks involve multiple competing
objectives ($m>1$), known as Multi-Objective Optimization (MOO).
These objectives can be represented as
\begin{equation}
f(x) = [f_1(x), f_2(x), \dots, f_m(x)].
\label{eq:moo_vector}
\end{equation}

Because the objectives typically trade off against one another, there is
generally no single solution that optimizes all $m$ objectives simultaneously.
Instead, the goal is to characterize the optimal tradeoff set, known as the
\emph{Pareto frontier}.
To evaluate the quality of candidate Pareto sets, a reference point
\[
r = [r_1, r_2, \dots, r_m]
\]
is defined, where $r_i$ is the minimum acceptable threshold for the $i$th
objective.

A widely used metric for assessing Pareto set quality is the \emph{Hypervolume}
(HV), which measures the $m$-dimensional volume of the region in objective
space dominated by the Pareto set $P = \{y_i\}_{i=1}^{k}$ and bounded by the
reference point $r$.
Formally,
\begin{equation}
HV(P,r) = \lambda \!\left( \bigcup_{y_i \in P} [\,r,\, y_i\,] \right),
\label{eq:hypervolume}
\end{equation}
where $\lambda$ denotes the Lebesgue measure, and $[\,r, y_i\,]$ represents the
hyperrectangle spanned between $r$ and the point $y_i$.

To optimize hypervolume gain in BO, the acquisition function commonly used is
the \emph{Expected Hypervolume Improvement} (EHVI):
\begin{equation}
EHVI(x) = \mathbb{E}\!\left[\big(HV(P \cup \{f(x)\}, r) - HV(P, r)\big)_{+}\right],
\label{eq:ehvi}
\end{equation}
where $(\cdot)_+$ denotes truncation at zero. This quantity measures the
expected increase in hypervolume resulting from evaluating a new candidate
point $x$.

When evaluating a batch of $q$ points simultaneously, the objective vector
becomes
\begin{equation}
F(X) = [\,f(x_1), \dots, f(x_q)\,].
\label{eq:batch_eval}
\end{equation}
In such settings, Monte Carlo methods are typically employed to estimate EHVI
due to the computational complexity of hypervolume calculations in higher
dimensions\cite{Daulton2020ehvi}.
In this work, we use the qLogEHVI acquisition function\cite{Ament2023logei},
which computes the logarithmically transformed EHVI objective in a numerically
stable way, yielding improved numerical properties and achieving state-of-the-art
multi-objective optimization performance, even compared to entropy-search-based approaches.
 
Multi-Objective Bayesian Optimization (MOBO) -- combining Gaussian Process regression with EHVI --
efficiently balances exploration and exploitation, enabling
the discovery of high-performing trade-offs within high-dimensional
design spaces.

\vspace{10pt}

\noindent{\textit{Gaussian Processes for Strength Modeling}}
\vspace{10pt}

We detail the model originally proposed by Ament et al.~(2023)\cite{Ament2023sustainable}.
Given the limitations of conventional machine learning models, 
we use a probabilistic model $\mathrm{strength}(x,t)$ that jointly captures the
dependence of compressive strength on both mixture composition $x$ and time
$t$. To enforce the physical condition that concrete has zero strength at the
time of casting, artificial zero-strength observations at $t=0$ are added for
both observed and randomly selected unobserved compositions.

Because strength development is highly nonlinear—rising rapidly at early ages
and plateauing at later times—a logarithmic transformation $t \rightarrow
\log t$ is applied to the time input.
This enables the use of stationary kernels and improves predictive
performance\cite{Pfeiffer2024bayesian}.
The kernel of the model combines a composition-independent, time-dependent
component $k_{\mathrm{time}}$ with a joint kernel over composition and time
$k_{\mathrm{joint}}$, yielding the composite kernel:
\begin{equation}
k\big((x,t),(x',t')\big)
= \alpha\, k_{\mathrm{time}}\!\left(\log t,\, \log t'\right)
+ \beta\, k_{\mathrm{joint}}\!\left((x,\log t),(x',\log t')\right),
\label{eq:gp_kernel}
\end{equation}
where $\alpha, \beta > 0$ are variance parameters learned via marginal
likelihood optimization\cite{Rasmussen2004gp}. An exponentiated quadratic kernel is
used for $k_{\mathrm{time}}$, while a Matérn~5/2 kernel with Automatic
Relevance Determination (ARD) is used for $k_{\mathrm{joint}}$. This structure
allows the model to learn general strength-development patterns while remaining
sensitive to compositional variations across mixtures.

\vspace{10pt}

\noindent{\textit{Model Performance Evaluation}}
\vspace{10pt}

The predictive performance of the trained GP model was evaluated using two
metrics: the Coefficient of Determination ($R^{2}$) and the Root-Mean-Square
Error (RMSE).
The $R^{2}$ value measures how well the independent variables explain the
variance in the dependent variable, with values closer to 1.0 indicating better
model agreement (reported as 0 if negative).
RMSE quantifies the average magnitude of the absolute error between predicted
and observed values.
The expressions for $R^{2}$ and RMSE are:

\begin{equation}
R^{2} = 1 -
\frac{\sum_{i=1}^{N} \big(y_{i} - \hat{y}_{i}\big)^{2}}
     {\sum_{i=1}^{N} \big(y_{i} - \bar{y}\big)^{2}},
\label{eq:R2}
\end{equation}

\begin{equation}
\mathrm{RMSE} =
\sqrt{\frac{1}{N}\sum_{i=1}^{N} \big(y_{i} - \hat{y}_{i}\big)^{2}},
\label{eq:rmse}
\end{equation}

where:

\begin{itemize}
\item $y_{i}$ denotes the observed output,
\item $\hat{y}_{i}$ denotes the predicted output,
\item $\bar{y}$ is the mean of the observed outputs, and
\item $N$ is the total number of observations.
\end{itemize}

\vspace{10pt}
\noindent\textbf{Materials and Experimental Methods}
\vspace{10pt}

\noindent{\textit{Materials}}
\vspace{10pt}

The materials used in this study were procured from a commercial supplier.
Portland Limestone Cement (PLC) was employed as the primary binder. Class C fly ashes, including C1 and C2 (7-day R3 heat release: 316~J/g), along with a Class F fly ash F1 (7-day R3 heat release: 209~J/g) and Grade 100 slag including S1, S2 and S3 (7-day R3 heat release: 488~J/g), were incorporated as Supplementary Cementitious Materials (SCMs) to partially replace PLC.
The chemical compositions of the binders were quantified using X-Ray
Fluorescence (XRF) in accordance with ASTM~D4326-13\cite{ASTM2021xrf} and are
reported in the Supplementary Information.
Particle Size Distributions (PSDs) of the binders were obtained using a laser
diffractometer, with results presented in the Supplementary Material.

\vspace{10pt}

\noindent{\textit{Mix Dataset Development}}
\vspace{10pt}

The mix development followed a closed-loop experimental design framework (Fig.~\ref{fig:Fig1}), combining
iterative laboratory optimization with GP-based model refinement.
The experimental dataset comprises 123 unique formulations: 69 mortar mixes and
54 concrete mixes, developed across six sequential phases.
Each phase systematically expanded the parameter space of binder composition,
Water-to-Binder (w/b) ratio, cement replacement level, High-Range Water Reducer
(HRWR) dosage, and curing temperature—all key parameters influencing hydration
kinetics, strength development, and environmental impact.

The range of mix design constituents for Dataset~1 is summarized in Table~2,
with the training mixes for Datasets~2--5 summarized in Tables~S2--S5 of the
Supplementary Information.
Fig.~\ref{fig:Fig5} graphically summarizes the compositional variability of the experimental
dataset and its influence on compressive strength.
The scatter plots in the upper panels depict how the 28-day compressive
strength ($f'_c$) varies as a function of individual mixture
parameters—including cement, fly ash, slag, aggregate fractions, HRWR dosage,
paste content, SCM replacement level, and w/b ratio.
Each subplot presents distinct performance trends: increases in binder content
and reductions in w/b ratio generally yield higher strengths, while elevated
SCM contents or higher HRWR dosages introduce variability due to differences in
pozzolanic reactivity and chemical dispersion efficiency.

The ternary plots in the lower panels represent the binder compositional
space—cement, fly ash, and slag—colored by compressive strength at 1, 5, and
28~days.
The color gradients illustrate the evolution of strength with hydration age,
highlighting the transition from clinker-dominated early-age strength (1~day)
to SCM-activated later-age performance (28~days).
These visualizations collectively demonstrate how cement and SCM proportions
jointly govern strength development trajectories across curing stages.

\begin{figure}[p]
    \centering
    \includegraphics[width=0.75\textwidth]{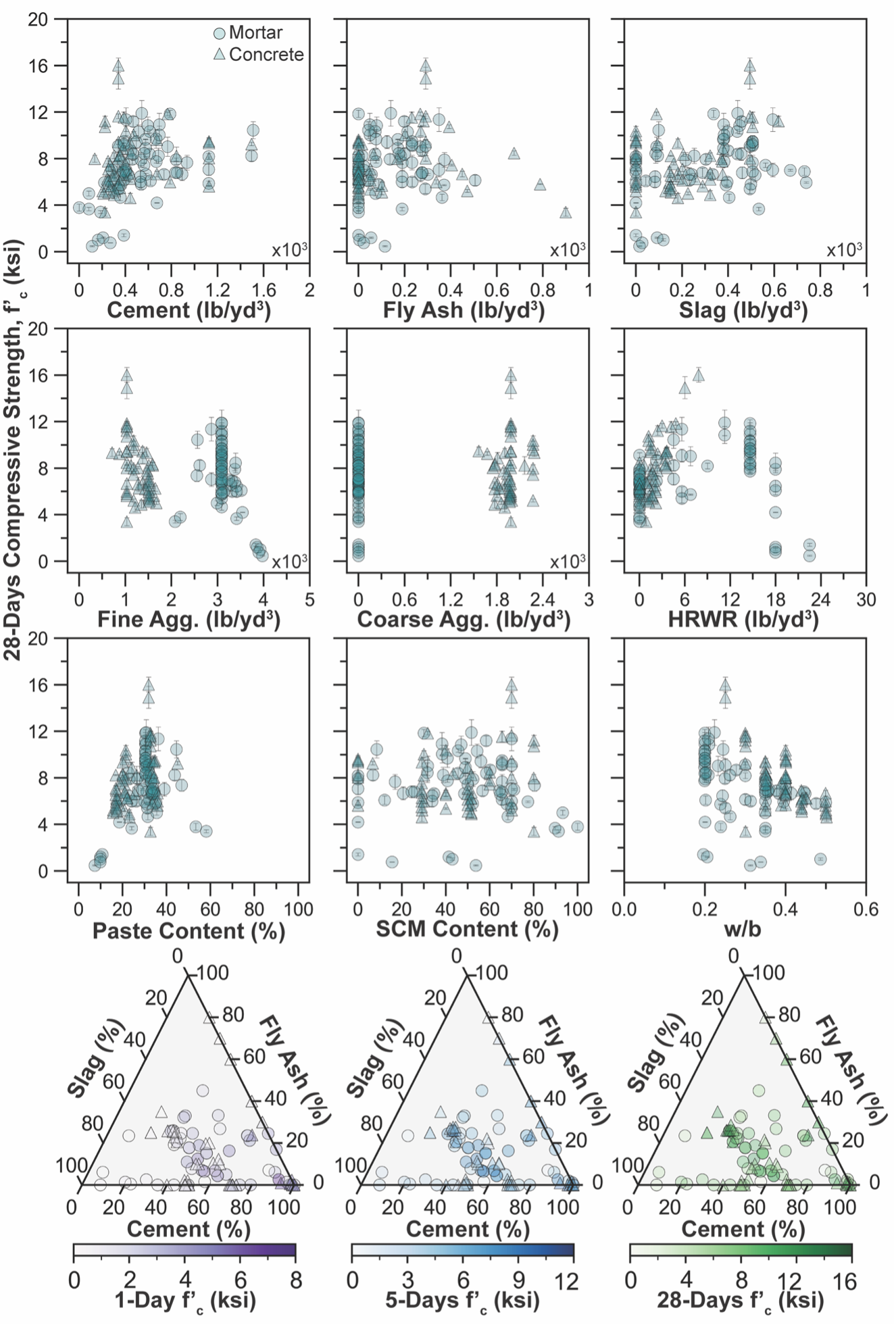}
    \caption{
\textbf{Compositional diversity and binder synergy governing strength development.}
Scatter plots show the distribution of all mortar and concrete mixtures, relating 28-day compressive strength ($f'_c$) to key mix parameters including cement, fly ash, slag, aggregates, HRWR dosage, paste content, SCM replacement, and water-to-binder (w/b) ratio. The broad, non-monotonic trends reflect the intentionally diverse design space spanning both high- and low-clinker systems, forming the basis for robust model training. Ternary diagrams depict binder phase distributions (cement--slag--fly ash) at 1, 5, and 28 days, with strength indicated by a continuous color gradient. Early-age strength concentrates in cement-rich compositions, whereas later-age high-strength regions shift toward balanced ternary blends, highlighting progressive SCM activation and synergistic hydration effects.
}
    \label{fig:Fig5}
\end{figure}


The dataset construction began with Phase~I, which focused on mortar-scale
experiments to explore a wide parameter range:

\begin{itemize}
    \item \textbf{Phase I} focused on mortar mixes ($n=69$) with a wide range
    of binder contents (240--2150~lb/yd$^3$), water-to-binder ratios
    (0.20--0.50), cement replacement levels (0--100\%), HRWR dosage
    (0--22.5~lb/yd$^3$), and curing temperatures (4.5--22~$^\circ$C).

    \item \textbf{Phase II} scaled up promising mortar formulations to concrete
    by adjusting the fine-to-coarse aggregate ratio (0.3--0.9).

    \item \textbf{Phase III} refined the scaled-up concrete mixes by
    modifying cement replacement levels (up to 80\%), water-to-binder ratios
    (0.30--0.40), and HRWR dosage (0--3~lb/yd$^3$), which were then
    reintroduced into the dataset.

    \item \textbf{Phase IV} expanded the design space by varying binder content
    (675--1125~lb/yd$^3$), water-to-binder ratios (0.35--0.50), SCM dosage
    (0--80\%), and HRWR dosage (0--1.3~lb/yd$^3$).

    \item \textbf{Phase V} utilized the GP model integrated with MOBO to
    generate new concrete formulations optimized for low GWP and high strength.
    These AI-recommended mixes featured binder contents of 625--755~lb/yd$^3$,
    cement replacement levels of 29--60\%, and HRWR contents of
    2.2--3.4~lb/yd$^3$. Following experimental validation, these mixes were
    incorporated into the training set.

    \item \textbf{Phase VI} fine-tuned the most promising AI-generated
    formulations by further adjusting binder content (440--760~lb/yd$^3$),
    cement replacement levels (29--52\%), water-to-binder ratios (0.39--0.50),
    HRWR dosage (0--3.4~lb/yd$^3$), and curing temperatures (10--22~$^\circ$C)
    to enhance model robustness.
\end{itemize}

For model evaluation, the dataset was split into an 80\% training set and a
20\% testing set.
Predictive performance was assessed by comparing GP model predictions on the
test set against experimentally measured strengths.

\begin{table}[H]
\centering
\small
\caption{Summary of Experimental Mix Development Phases and Key Variable
Adjustments for Training Set 1.}
\vspace{10pt}

\begin{tabular}{l l l p{4.5cm}}
\toprule
\textbf{Phase} & \textbf{Mixes} & \textbf{Type} & \textbf{Parameters Modified} \\
\toprule

P I  & 69 (M1--M69) & Mortar
& Binder content, w/b ratio, cement replacement, HRWR dosage, curing temp. \\
\midrule

P II & 7 (C1--C4, C6--C8) & Concrete
& Fine-to-coarse ratio \\
\midrule

P III & 5 (C10--C14) & Concrete
& SCM dosage, w/b ratio, HRWR content \\
\midrule

P IV & 10 (C16--C20, C22--C25) & Concrete
& Binder content, w/b ratio, SCM dosage, HRWR content \\
\midrule

P V  & 4 (C28--C31) & Concrete
& Binder content, cement replacement, HRWR content \\
\midrule

P VI & 16 (C33--C53) & Concrete
& Binder content, cement replacement, w/b ratio, HRWR dosage, curing temp. \\
\bottomrule
\end{tabular}

\label{tab:phases}
\end{table}

\vspace{10pt}

\noindent{\textit{Compressive Strength Testing}}
\vspace{10pt}

Compressive strength was determined for both mortar and concrete specimens.
Mortar cubes (2~$\times$~2~in.) were tested in accordance with
ASTM~C109\cite{ASTM2024mortar}, while concrete cylinders (4~$\times$~8~in.) were
prepared and tested according to ASTM~C39\cite{ASTM2023cylinder}.
The mortar samples were tested at four ages (1, 3, 5, and 28~days), whereas the
concrete samples were tested at five ages (1, 3, 5, 14, and 28~days).
The 1-day tests were conducted immediately after demolding, and all specimens
were subsequently stored in a curing chamber maintained at 98\% relative
humidity until the designated testing ages.

\vspace{10pt}

\noindent{\textit{Global Warming Potential Assessment}}
\vspace{10pt}

The Global Warming Potential (GWP) of the developed concrete mixtures was
assessed on a cradle-to-gate basis using the following expression:

\begin{equation}
GWP(z_i) = g_m^{\top} z_i + g_t^{\top} z_i + g_p^\top z_i,
\label{eq:gwp}
\end{equation}

where $g_m$, $g_t$, and $g_p$ represent the Greenhouse Gas (GHG) emissions
associated with raw material production, transportation, and concrete
production, respectively. The vector $z_i$ denotes the quantity of each
constituent used in mix $i$\cite{Pfeiffer2024bayesian,Kim2022openconcrete}.

GWP calculations were performed using \textit{OpenConcrete}, an online life
cycle assessment tool that adopts a process-based LCA methodology.
The tool models emissions from each stage of concrete production—including raw
material extraction, grinding, and transportation to the batch plant.
Supplementary Fig.~S1 illustrates the life cycle stages of major concrete
constituents.
OpenConcrete allows users to specify custom mix compositions, transport
distances, and regional energy mixes, enabling context-sensitive impact
estimation.

To ensure consistency and comparability, all GWP values were normalized per
cubic meter of concrete\cite{ASTM2023cylinder}.

\section*{Acknowledgements}

This work was supported by Meta Platforms, Inc. through a research
collaboration with the University of Illinois Urbana-Champaign.
Experimental activities were conducted in the Newmark Civil Engineering
Laboratory, Department of Civil and Environmental Engineering, University of
Illinois Urbana-Champaign.
The authors gratefully acknowledge the contributions of Xiaorui Gu, Jocelyn
Pytel, Zeeshan Imtiyaz Ahmed Ranginwala, Hayden Thyu Bui, Charlie Knight, Ayush
Jayakannan, Thai Vu Tran, and Ji~Won Park for their assistance with specimen
preparation, mixing, and testing.
Andrew Witte is sincerely acknowledged for generating and sharing the mortar
dataset that forms an integral part of this study.

\section*{Author Contributions}

B.B.\ and M.A.I.\ conducted the laboratory experiments, performed data
analyses, generated visualizations, and prepared the initial manuscript draft.
S.A.\ developed the Gaussian Process model and multi-objective optimization
framework that guided, analyzed, and optimized the experiments.
J.K.\ and N.G.\ jointly supervised the project and provided technical direction
throughout the study.

\bibliography{references}

\end{document}